\makeatletter
\def\input@path{{styles/}}
\makeatother
\documentclass[runningheads]{styles/llncs}

\makeatletter
\let\NAT@parse\undefined
\makeatother

\usepackage{hyperref}
\usepackage[utf8]{inputenc}
\usepackage{authblk}
\usepackage[ruled, linesnumbered]{algorithm2e}
\usepackage{algorithmicx}

\usepackage{marginnote}
\usepackage{amsfonts}
\usepackage{amsmath}
\usepackage{graphicx}
\usepackage[usenames,dvipsnames,table,xcdraw]{xcolor}
\usepackage{xspace}
\usepackage{caption}
\usepackage[table]{xcolor}
\usepackage{multirow}

\usepackage[symbol]{footmisc}

\newcommand{\Term}[1]{\textsf{#1}}

\usepackage{subcaption}

\usepackage[nameinlink]{cleveref}

\definecolor{almostblack}{rgb}{0, 0, 0.3}
\providecommand{\emphw}[1]{{\textcolor{almostblack}{\emph{#1}}}}%

\newcommand{\FK}{F_K}

\hypersetup{%
      breaklinks,%
      colorlinks=true,%
      urlcolor=[rgb]{0.25,0.0,0.0},%
      linkcolor=[rgb]{0.5,0.0,0.0},%
      citecolor=[rgb]{0,0.2,0.445},%
      filecolor=[rgb]{0,0,0.4},
      anchorcolor=[rgb]={0.0,0.1,0.2}%
}

\newcommand{\cspace}{\ensuremath{\mathcal{C}_{space}}}

\newcommand{\R}{\mathbb{R}}

\newcommand{\DOF}{\Term{DOF}\xspace}
\newcommand{\RRT}{\Term{RRT}\xspace}
\newcommand{\RRTstar}{\Term{RRT\ensuremath{^*}}\xspace}
\newcommand{\PRM}{\Term{PRM}\xspace}
\newcommand{\SBMP}{\Term{SBMP}\xspace}
\newcommand{\SIPP}{\Term{SIPP}\xspace}
\newcommand{\AABB}{\Term{AABB}\xspace}
\newcommand{\LazyPRM}{\Term{LazyPRM}\xspace}

\renewcommand{\th}{th\xspace}

\newcommand{\etal}{\textit{et~al.}\xspace}

\newcommand{\si}[1]{#1}

\newcommand{\HLink}[2]{\hyperref[#2]{#1~\ref*{#2}}}
\newcommand{\figlab}[1]{\label{fig:#1}}

\author{
Stav Ashur\and
Maria Lusardi$^*$ \and
Marta Markowicz$^*$ \and
James Motes \and
Marco Morales \and
Sariel Har-Peled \and
Nancy M. Amato
}
\footnotetext[1]{Equal contribution}
\authorrunning{S. Ashur et al.}
\titlerunning{SPITE}

\institute{Department of Computer Science, University of Illinois, 201 N. Goodwin Avenue, Urbana, IL 61801, USA}

\title{
SPITE: Simple Polyhedral Intersection Techniques for modified Environments
}

\date{March 2024}

\begin{document}

\maketitle

\begin{abstract}
    Motion planning in modified environments is a challenging task, as it
    compounds the innate difficulty of the motion planning problem with a changing
    environment.
    This renders some algorithmic methods such as probabilistic 
    roadmaps less viable, as nodes and edges may become invalid as a result of these
    changes.
    In this paper, we present a method of transforming any configuration 
    space graph, such as a roadmap, to a dynamic data structure capable of updating
    the validity of its nodes and edges in response to discrete changes in obstacle positions.
    We use methods from computational geometry to compute 3D swept volume
    approximations of configuration space points and curves to achieve 10-40
    percent faster updates and up to 60 percent faster motion planning queries than previous algorithms while requiring a
    significantly shorter pre-processing phase, requiring minutes instead of
    hours needed by the competing method to achieve somewhat similar update times.
\end{abstract}


\section{Introduction}

\begin{figure}[t]
    \centering
    \includegraphics[width=0.8\linewidth]%
    {\si{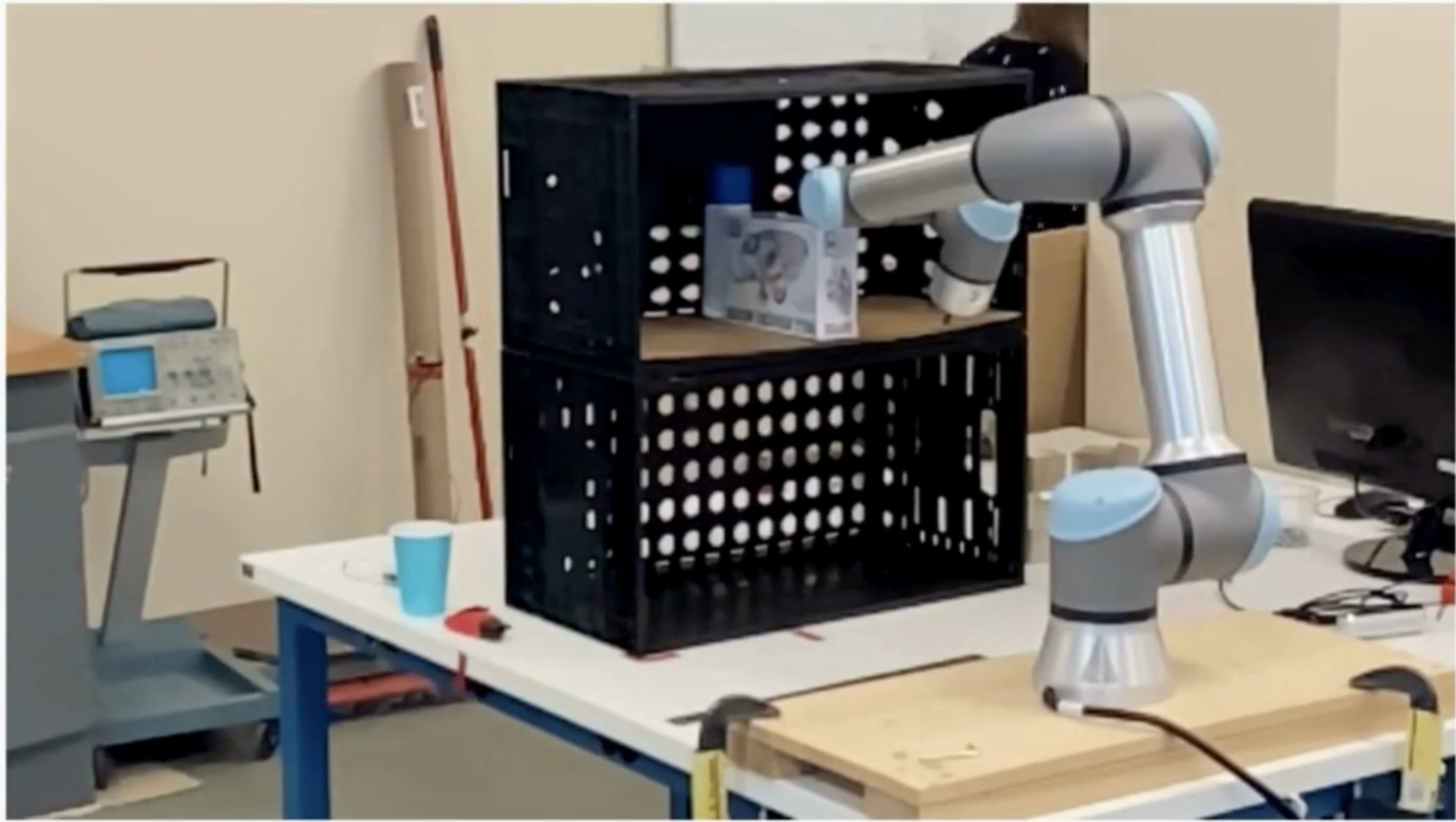}}

    \caption{An image of the 5\DOF UR5e robot used in physical experiments to test the validity of paths produced in simulation. Shelf environments are likely to be modified as stored items are added, removed, or placed in new locations.}
    \label{fig:ur5}
\end{figure}

In the motion planning problem, we are given a robot $r$ and an
environment $E = \{B, O\}$ (aka \emph{workspace}) which is composed of a bounding
box $B$, a set of obstacles $O$, and two configurations $s$ and $t$
of $r$ in $E$, and we are tasked with deciding whether there exists a valid,
i.e. collision free, sequence of movements taking $r$ from configuration
$s$ to $t$.

A common approach in designing motion planning algorithms is to reduce
the problem to finding a curve in the space of valid configurations,
a subset of the implicitly defined \emph{configuration space}
(\emph{$C$-space})~\cite{lw-apcfpapo-79} of $r$, with the points
corresponding to configurations $s$ and $t$ as its endpoints. These
algorithms often build a geometric graph in the $C$-space called a \emph{roadmap} 
whose nodes and edges
represent feasible configurations and valid continuous motions of $r$ in $E$
respectively~\cite{kslo-prpp-96}. These representations often assume an environment in which obstacle positions never change.

\begin{figure}[t]
    \centering
    \hfill
    \begin{subfigure}{0.35\linewidth}
        \includegraphics[page=1, width=\linewidth]{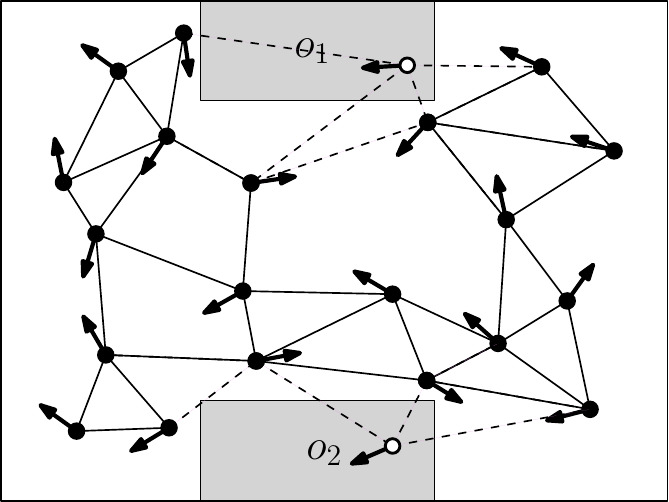}
        \caption{Before $o_2$ updated position}
        \label{fig:obstacle:move:before}
    \end{subfigure}
    \hfill
    \begin{subfigure}{0.35\linewidth}
        \includegraphics[page=2, width=\linewidth]{figs/obstacle_move.pdf}
        \caption{After $o_2$ updated position}
        \label{fig:obstacle:move:after}
    \end{subfigure}
    \hspace{1cm}
    \caption{A modified environment $E=\left\{B, \{o_1, o_2\}\right\}$
       with an overlaid roadmap for a small robot with two
       translational \DOF{}s and one rotational \DOF. The valid and
       invalid nodes of the graph are depicted as configurations with
       black and white centered points respectively, and valid and
       invalid edges are depicted as full and dashed segments
       respectively. The obstacle $o_2$ moves from its first position
       seen in \ref{fig:obstacle:move:before} to a new
       one seen in \ref{fig:obstacle:move:after}, an event that
       changes the validity of several nodes and edges.}
    \label{fig:obstacle:move}
\end{figure}

In this paper, we tackle the problem of enhancing such roadmaps to account 
for the possibility of discrete changes in the
obstacles' positions. These changes occur in multi-agent task and
motion planning problems (i.e., multi-robot and human-robot
collaborations), where agents can move objects, or in methods which
seek to re-utilize plans generated earlier.
See Figure \ref{fig:obstacle:move} for an illustration.

The main challenge is to update the validity of nodes and edges in the roadmap
(corresponding to points and curves in $C$-space) given changes in
the workspace, as the effects of such changes on the $C$-space are not
well understood. Intuitively, by continuity, small changes in object
positions should result in local effects in the roadmap, as it is a
geometric graph. Thus, one can hope that a relatively efficient graph
data structure can help with this task.

\paragraph*{\textbf{Contribution}}
We present a supplemental algorithm for roadmaps that can
quickly find the set of nodes and edges whose validity status was
altered due to a change in the workspace. While our contribution is not a motion planning algorithm, it is designed for multi query motion planning scenarios and enables quick computation of motion planning  queries in a single environment for a known robot. Any graph-based \SBMP algorithm can be paired with our method in order to generate an augmented roadmap that can maintain its functionality in the presence of workspace changes using less pre-processing time and faster updates than the previously best known method.

We achieve this improvement with a second contribution, an approach to 3D workspace
volume approximations which enables us to store approximated robot configurations and swept volumes in a $kD$-tree like data structure which we use for fast hierarchical collision checking. 
This enables us to obtain a set of nodes and edges which may be affected by some modification made to the environment, and either invalidate or re-validate them.

\paragraph*{\textbf{Evaluation}}

We compare our method with what is, to the best of our knowledge, the only method that produces a dynamic
roadmap capable of updates in the presence of workspace changes \cite{lh-frtppce-02,km-mpdr-04} 
Since there is no extant code from either paper, we have implemented the algorithm from \cite{km-mpdr-04} ourselves, and tested
our method against it both by comparing query times with a wide range of
obstacles, and by comparing the runtimes of sequences of motion planning
problems in a changing environment.

Similarly to \cite{km-mpdr-04}, we also tested the
runtime of motion planning queries computed by applying our method to a roadmap against single-query motion planning algorithms often used for non-static scenarios.
Like \cite{km-mpdr-04}, we use \RRT \cite{l-rrtntpp-1998} for these comparisons, and, for reasons later explained, chose to also include the \LazyPRM algorithm \cite{bk-ppulp-00} in our experiments.

Our experiments confirm that our data structure performs faster update
operations than the only previous method, with average runtimes 10-40 percent faster depending on the size and shape of the modified obstacle (see Section
~\ref{sec:update:experiments} for details and Table~\ref{table:Update:times}
for results) and enables an augmented roadmap to solve motion planning queries faster than both single query algorithms and roadmaps utilizing the previous method by as much as $\sim 60$ percent (see Section~\ref{sec:motion:planning:experiments} for
details and Tables~\ref{table:motion:planning} and~\ref{table:ur5} for results).


\section{Related Work}
\label{sec:related:work}

In this section, we present the relevant body of work published on
the problem and related subjects. We start by giving a brief review of
some of the seminal work on sampling based motion planning, followed
by a more in depth review of such work specifically designed for
changing environments. We divide the latter part into two, \LazyPRM
adjacent work, and \RRT adjacent work. Note that \LazyPRM and \RRT are
the methods we use to demonstrate our contribution in Section
\ref{sec:motion:planning:experiments}.


\subsection{Sampling-based motion planning}

Sampling based motion planning (\SBMP) is a motion planning approach
in which $C$-space graphs are built using random sampling of nodes which
are connected by edges computed by motion-planning primitives called
\emph{local planners}. This addition of randomness has been found
effective in addressing the problem's intractability that expresses
itself as a $C$-space of possibly exponential complexity in the size of
the input.

Probabilistic Roadmaps (\PRM), introduced by Kavraki \etal
\cite{kslo-prpp-96}, are a family of \SBMP algorithms that usually
randomly sample node configurations, connect each configuration to
some set of nearby nodes to form a graph (aka \emph{roadmap}) containing representations of feasible paths. The roadmap is used to
solve the motion planning problem by adding the start $s$ and the goal $t$ as new nodes
of the graph and returning a graph $(s,t)$-path if one exists.

\PRM{}s are usually used for the multi-query variant of the motion
planning problem in which we are required to solve a set of motion
planning problems for a single type of robot in a static workspace, as
any change to either of these components results in changes to the
implicit $C$-space and might render some of the graph's nodes and edges
invalid. Many \PRM variants exist and have been found to be applicable
for a wide range of motion planning problems. See~\cite{ock-sbmpcr-23}
and references therein.

The \RRT algorithm introduced by LaValle \cite{l-rrtntpp-1998} is a
single query motion planning algorithm that expands a tree graph in the
$C$-space by choosing a random point in the space
and extending an edge towards that point from the nearest tree
node. The \RRT algorithm and its many variants, see \cite{es-sbmpr-14}
and references therein, are widely used both as stand-alone single query
algorithms, and as a building block in many more complex methods, some
of which we discuss below.


\subsection{Planning in Modified Environments}

The term ``Modified environments'' can be interpreted in two distinct
ways. In this paper the changing nature of the environment is
manifested in discrete changes to the location of obstacles, while
other motion planning algorithms operate in the presence of a temporal
dimension, meaning that certain obstacles have either known or
estimated trajectories over time, and the algorithm must take these into
account when computing the robot's path. For the rest of this paper,
this section excluded, we will use the adjective ``modified'' to mean the former.

\paragraph*{\textbf{Dynamic Roadmaps}}

The papers most closely related to ours are by Leven and
Hutchinson~\cite{lh-frtppce-02} and Kallmann and
Mataric~\cite{km-mpdr-04}, where the latter built upon ideas from the
former. Most notably, both papers approach the $C$-space - workspace
relationship by partitioning the workspace using a fixed resolution
grid, and maintaining lists of cell-node and cell-edge
incidences. Despite the strong connection between the papers, the two focus on different aspects of the problem. The first paper
(\cite{lh-frtppce-02}) utilizes the similarity of lists of adjacent
cells, where the best definition of ``adjacency'' is one of the
paper's contributions, to create efficient representations of these incidence
lists. The second (\cite{km-mpdr-04}) introduces a new update
operation which relies on approximating the moved obstacle by its
bounding box and focuses on the runtime when compared against \RRT.
Our contribution is much more closely aligned with that
of~\cite{km-mpdr-04}, but where they used unions of uniform
axis-aligned cubic grid cells in order to approximate the swept
volumes, we use arbitrarily oriented and shaped \emph{cigars}
(see Section \ref{sec:preliminaries}) to allow quick
intersection checks and constant size space complexity per
node/edge. Furthermore, where they used a uniform grid partition of
workspace (even though both they and ~\cite{lh-frtppce-02} mention the
possibility of using an octree) we use a hierarchical, input sensitive
decomposition. See Subsection~\ref{sec:cigar-tree}.

\paragraph*{\textbf{Lazy Evaluation}}
Bohlin and Kavraki's \LazyPRM \cite{bk-ppulp-00} is a well known
single query variant of the \PRM algorithm. The lazy algorithm creates
a roadmap without validating its edges, the part of \PRM which usually
requires the most runtime by an order of magnitude as a single edge
can contain thousands of configurations that require validation. Some
variants do not validate the nodes as well~\cite{bk-ararpp-01,bk-ppulp-00}. Given a motion
planning query, the algorithm lazily connects the start and the goal
to the roadmap and tries validating only the edges that lie on the
shortest $(s,t)$-path in the roadmap. While this is not explicitly an
algorithm meant for modified environments its ``laziness'' means that
it can operate in modified environments (as we describe them) by simply
ignoring all past information after each query.

\LazyPRM has also inspired other algorithms meant specifically for
changing environments. Jaillet and Sim\`{e}on~\cite{js-apbmpdce-04}
gave a \PRM based algorithm that computes a roadmap for the environment
while considering only the static obstacles, and lazily evaluating a
path by collision checking it only against dynamic obstacles that have
moved in the vicinity of an $(s,t)$-path found in the roadmap. An
important thing to notice is that if all of the obstacles are dynamic
this algorithm degenerates to a variant of \LazyPRM (as it does employ
several other heuristics not found in \cite{bk-ppulp-00}).

Hartmann \etal \cite{hot-eppmpparve-23} describe a single query lazy
motion planning algorithm for environments with \emph{movable} objects
that reuses computations by dividing validity checking efforts into
several parts, and, given a query, validating the path by validating
it against movable objects that have recently moved. Additionally,
they give a \PRM variant that, among other contributions, validates the
robots' collisions with itself and with static obstacles only when
expanding the roadmap, and, much like \cite{js-apbmpdce-04}, reuses
that information when validating a path.

\paragraph*{\textbf{Rapidly-exploring Random Tree (\RRT) Algorithms}}

\RRT is used in motion planning for changing environments due to its
ability to quickly explore an unknown $C$-space. This property is
extremely useful in these settings since a change to an obstacle's
position changes the landscape of the $C$-space and possibly invalidates
some of a roadmap's nodes and edges. An exploration process can
then be employed in order to find a new path in a region where a path
once existed~\cite{js-apbmpdce-04}.

Multipartite \RRT \cite{zkb-mrrt-2007} updates the validity of edges
that have been obstructed by moved obstacles and stores the
resulting disconnected subtrees in a cache. The root nodes of these
subtrees are used as part of subsequent sampling. RT-\RRTstar
\cite{naderi2015rt} grows an \RRTstar\cite{kf-isbaomp-10} tree, an
asymptotically optimal variant of \RRT. When the obstacle moves, it
rewires the tree to not include any invalidated edges, and when the
agent moves, the root is moved and the tree rewired around it to allow
for multiple queries. In order to do this rewiring the paper searches
adjacent cells with a simple grid-based spatial indexing.

Instead of taking a reactive planning approach, the workspace can be
extended by a temporal dimension to incorporate known or predicted
obstacle trajectories. \SIPP \cite{pl-ssippfde-11} utilizes safe
temporal intervals in a discretized space, where valid paths are found
by using A* to connect cells during safe intervals. ST-\RRTstar
\cite{grothe2022st} plans over continuous space with an added temporal
dimension using random sampling. It uses a number of approaches to optimize
sampling in both the spatial and temporal dimensions, such as
conditional and weighted sampling. These planning methods are thus
able to plan into the future around obstacle trajectories over time.

Note that the methods mentioned in the previous paragraph solve the single-query motion planning in a dynamic setting where obstacles move during the
query phase. We therefore do not compare our data structure to these methods
as they use extra computational resources for responding to online changes
unlike single query algorithms for static environments.


\section{Preliminaries}

\label{sec:preliminaries}

Let $r$ be a robot, $E = (B, O = \{o_i: i=1,\ldots, m\} )$ be a 3D
environment, $\cspace$ be the implicit $C$-space defined by
$r$ and $E$, and $G$ be a geometric graph in $\cspace$, i.e. a roadmap
whose nodes and edges are points (configurations) and segments in
$\cspace$ respectively.

\paragraph{Problem Definition}

Given $r, E, O,$ and $ G$ as an input, compute a dynamic graph $H$
capable of performing 6\DOF transformation updates of an object's
location in workspace $o_i \mapsto R\cdot o_i + T$. At the end of the
update operation all valid nodes and edges must be labeled as
valid, and all infeasible nodes and edges must be labeled as invalid.

Let $\FK : \cspace \longrightarrow P\left(\R^3\right)$ be the
\emph{forward kinematics function} that, given a configuration
$c\in \cspace$, returns the set of all points in $\R^3$ that are
occupied by $r$ assuming the configuration $c$. With a slight abuse of
notation we use $\FK(C)$ for subsets $C\subseteq \cspace$ to mean
$\bigcup_{c\in C}\FK(c)$, and call the resulting 3D set the
\emph{swept volume} of $C$.

A node or an edge of $G$ is \emphw{invalid} if its swept volume
intersects any obstacle $o_i$, and otherwise it is \emphw{valid}. Note
that this definition does not include robot self collision, kinodynamic
constraints, and out-of-boundary constraints. This is suitable for our
purposes as we only deal with movements of obstacles within $B$ and their
effect on $G$. This is not an issue because any of the aforementioned
conditions can be pre-computed once when computing $G$ and have no further
bearing on the problem. This simplification of the validity conditions
allows us to reduce validity checking to 3D shape intersections which can
be done very efficiently under some assumptions we consider to be reasonable.
Our assumptions are as follows:

\begin{enumerate}
    \item $r$ can be approximated as a small set $b^r_1,...,b^r_k$ of
    convex polyhedrons.

    \item Every object $o_i\in O$ can be approximated as a small set
    $b^{o_i}_1,..,b^{o_i}_l$ of convex polyhedrons.

    \item The approximations mentioned above are either given as part
    of the input or can be efficiently computed.
\end{enumerate}


\section{Method}
\label{sec:method}

As mentioned previously, in many high dimensional motion planning
problems the relationship between the $C$-space and the workspace is
neither easy to understand nor to compute. As such, we use rough
approximations of the mapping between points and segments in the
high-dimensional $C$-space to 3D volumes in workspace, and utilize
common computational geometry tools in order to perform roadmap
update operations.  The shapes we are required to approximate are
swept volumes of points and segments in $\cspace$, which, under our
assumptions, are small sets of convex polyhedrons or linear motions
(with rotations) of such polyhedrons. We approximate these polyhedrons
using \emph{cigars}.

A cigar, also called a capped cylinder, is the
Minkowski sum $s\oplus b$ of a segment and a ball, that is
$\bigcup_{p\in s}b(p)$ where $b(p)$ is a translated copy of the ball
$b$ centered at $p$, and it is our ``weapon of choice'' for three main
reasons; $a)$ Cigars are very similar to ellipsoids which have been
long known to provide decent approximations for high dimensional
convex bodies \cite{j-episc-48}, $b)$ Linear motions of convex bodies
tend to create long thin swept volumes which can be easily bounded by
a cigar, and $c)$ Distance and intersection computations involving
cigars can be quickly approximated by viewing them as 3D segments with
an associated radius.


In the following subsections, we describe the method by which we
compute bounding cigars of swept volumes and create a data structure
capable of storing a set of cigars and answering intersection queries
of said set with simple shapes.


\subsection{Swept Volume Approximation}
\label{sec:swept-volume-approx}

Our assumptions immediately imply another assumption of many \SBMP
algorithms, which is that continuous motion can be sufficiently
approximated by a discrete set of intermediate configurations.
Given an edge $(p,q) \in G$, we first construct a point cloud $P$ containing
the vertices of $\FK(c)$ for every intermediate $c$ of $(p,q)$. We then
use an algorithm for approximating the optimal volume oriented bounding box of a point cloud~\cite{bhp-eamvbbpstd-01}, and construct our cylinder from the resulted box.
Note that rather then using the box itself as the 3D representation of the swept volume, we opted for using the smallest cigar containing the box for simpler and quicker intersection queries.

\begin{figure}[t]
    \centering
    \begin{subfigure}{0.23\linewidth}
        \includegraphics[page=5, width=\linewidth]{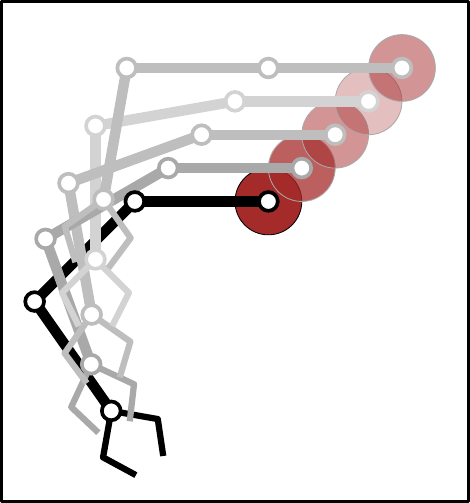}
        \caption{Intermediates composing an edge of a roadmap (planar manipulator)}
        \figlab{cigar:creation:intermediates}
    \end{subfigure}
    \begin{subfigure}{0.23\linewidth}
        \includegraphics[page=7, width=\linewidth]{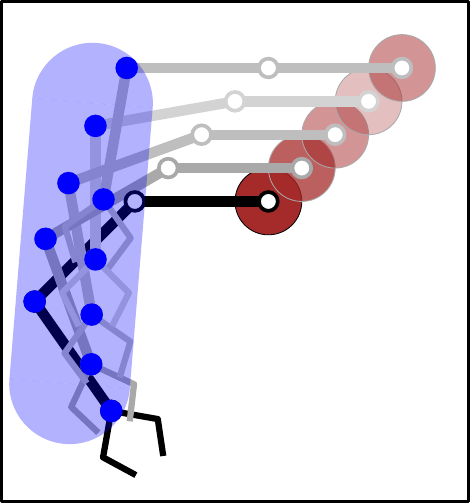}
        \caption{The swept volume approximation of the 3rd link of the
           robot}%
        \figlab{cigar:creation:cigar}
    \end{subfigure}
    \begin{subfigure}{0.23\linewidth}
        \includegraphics[page=11, width=\linewidth]{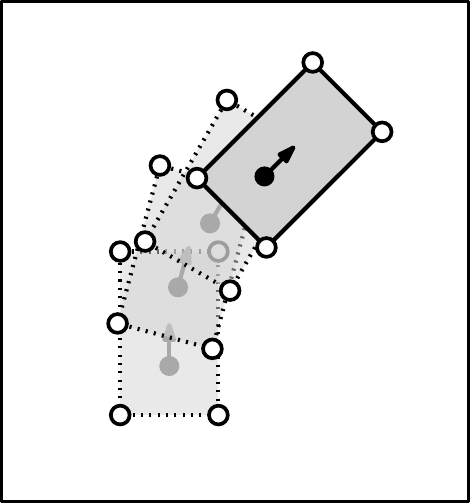}
        \caption{Intermediates for an edge of a roadmap (planar mobile robot)}
        \figlab{cigar:creation:intermediates}
    \end{subfigure}
    \begin{subfigure}{0.23\linewidth}
        \includegraphics[page=12, width=\linewidth]{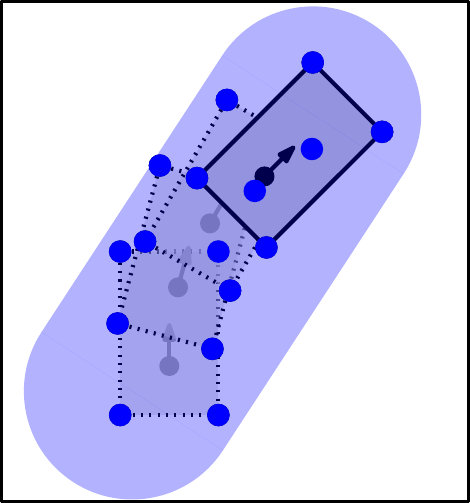}
        \caption{The swept volume approximation of the robot}%
        \vspace{0.39cm}
        \figlab{cigar:creation:cigar}
    \end{subfigure}
    \hspace{1cm}

    \caption{In (a) we illustrate a set of configurations of a mobile
       manipulator with 2 translational \DOF{}s and 3 angular joints
       composing an edge in some roadmap, and in (b) we see the
       point cloud and the cigar corresponding to the 3rd link of the
       robot. In (c) and (d) a similar process can be seen for a simple 3\DOF{} planar robot.}
    \figlab{cigar:creation}
\end{figure}


\subsection{Cigar Tree}
\label{sec:cigar-tree}

We store the cigars in an axis-aligned bounding box tree (\AABB tree),
a data structure resembling a $kD$-tree in which every node is
associated with an axis-aligned bounding box and the set of objects it
contains, and the node's children are constructed by splitting
the bounding box perpendicularly to one of the axes and using some
tie-breaker for objects intersecting the boxes of more than one child.
Our implementation actually uses a ternary tree with all of the objects
intersecting both
sides stored in a separate child. Such geometric data structures may
be problematic for certain inputs, for example a set of long segments
with endpoints close to antipodal points of the scene's bounding box
cannot be easily separated along any of the 3 axes. This is,
however, mostly irrelevant in our case due to the reasons described
earlier in this section.  Our data structure answers intersection
queries in a straightforward \AABB way. Given some query polyhedron
$Q$, we start at the root and recurse on the children whose bounding
box intersects $Q$, and, upon reaching a leaf of the tree, we check for
intersections between the small (constant) number of cigars stored in
that leaf and $Q$ and append those that intersect to the output list.

Using the tools described above in conjunction with a simple data
structure that stores for every obstacle $o_i$ the list of edges and
nodes it is currently invalidating we get a quick update operation
described in Algorithm \ref{alg:update_rm}. The obstacle's polyhedron
or approximating polyhedron is passed as an argument to the tree's
intersection query resulting in a list of nodes and edges that need to
be re-examined. This is followed by a re-examination of nodes and edges
that were previously in collision with $o_i$.

\begin{algorithm}[t]
    \caption{Roadmap Update}
    \label{alg:update_rm}
    \SetAlgoLined
    \SetKwInOut{Input}{input}
    \Input{Obstacle $o$, Transformation $t$}
    $o$.Update{}Location($t$)\\
    $invC \gets Tree$.Get{Intersecting}Objects($o$)\\
    \For{$c \in invC$}{
       \If{CD($c$,$o$) == blocked}{
          $c$.invalidate()\Comment{Invalidating nodes and edges}\\
          $c$.intersectionList.add($o$)
       }
    }
    $revC \gets Tree$.Get{Intersected}Volumes($o$)\\
    \For{$c \in revC$}{
       \If{$CD($c$,$o$) == free$}{
          $c$.intersectionList.remove($o$)\\
          \If{$c$.intersectionList.IsEmpty()}
          {$c$.validate()\Comment{Re-validating nodes and edges}}
       }
    }
\end{algorithm}

Some experiments not included in this paper actually show that the using
the tree, as opposed to, say, a simple list of cigars, has very small
effect on the runtime of graph update operations. This is true even though
it is exponentially faster then linear scanning, and it is due to the
runtime of the collision detection calls required to validate nodes and
edges dominating the overall resulted runtime. 


\section{Experiments}
\label{sec:experiments}

With our experiments, we aim to validate our claims that 
1) our approach results in faster update times for a dynamic roadmap in the presence of workspace changes, 
and 2) this faster update time of the dynamic roadmap enables faster motion planning in multi-query scenarios than single query algorithms traditionally used for non-static scenarios.

\subsection{Experimental Setup and Implementation Details}

In order to demonstrate our claims we run two sets of experiments.

In the first set, we directly measure the improvement in the roadmap update
time against a brute-force benchmark and against an implementation of the
method described in~\cite{km-mpdr-04}, which, as mentioned in
Section~\ref{sec:related:work}, is the only method for $C$-space
graph updates for modified environments we were able to find in the literature.
We refer to this algorithm as the \emph{grid method}. Note that we have
implemented the grid method ourselves based on the paper.

In the second set, we use our data structure as part of a multi-query motion
planning algorithm and compare its performance against the grid
method, \RRT, and \LazyPRM. These experiments capture the applicability of
our data structure in lieu of the results showcased by the first experiment,
and demonstrates its ability to allow fast pathfinding operations in
$C$-space in a modified setting for both mobile robots and manipulators. 
The grid method was previously only compared to \RRT, but we have decided to
compare our method against \LazyPRM as well since \LazyPRM is not only a fast
single query \SBMP algorithm, but one that can take an unvalidated roadmap as
an input which may considerably improve its performance, e.g., when the
provided roadmap contains a valid path. 

Our method and the grid method only maintain and update the
validities of the roadmap nodes and edges when obstacles are moved,
and so in order to fairly compare the runtime of motion planning
queries against single-query methods we measure the roadmap update
time in addition to the time needed for the roadmap to answer the 
motion planning query. We stress that our method does not
assist in pathfinding or the creation of better roadmaps, 
only adjusts such graphs to allow them to operate on modified environments. For this reason we
use an input graph guaranteed to contains a solution, and the same
graph is provided to all methods for a fair comparison. This choice
of roadmap demonstrates that, given an appropriate graph to build
upon, our method provides a better solution for multiple
queries in a modified environment than using the grid method or single
query algorithms. 

\subsubsection{SPITE Implementation}


Our method uses two computational geometry components for its purpose, one for approximating 3D volumes with cigars, and one for storing the cigars and answering intersection queries.
Since the runtimes related to the cigar tree operations have almost no impact on the overall runtime, as they are smaller than those required for full collision checks by orders of magnitude, we did not bother to optimize them.

The approach for computing the cigars employs an algorithm for constructing an approximately optimal oriented bounding box~\cite{bhp-eamvbbpstd-01} for which we have chosen the approximation factor $0.1$.

We determine the set of cigars intersecting an obstacle, and by extension, the set of nodes and edges possibly invalidated by a modification to the workspace, by querying the cigar tree with the obstacle's axis aligned bounding box. The set of previously invalid nodes and edges that need to be re-checked is compiled by storing an incidence list of obstacles and nodes/edges, and all fine-grained collision checking is done by the collision detection function.

\subsubsection{Grid Method Implementation}
The grid method partitions the workspace into uniform cubes of some pre-determined
side length and computes for each cell the set of nodes corresponding to 
configurations intersecting that cell and edges corresponding to movements whose
swept volume intersects it. This can be viewed as approximating the 3D volumes occupied
by robot configurations and movements by a subset of axis-aligned uniform size cubes.
Given an update in an obstacle's position the method computes the axis-aligned bounding
box of the obstacle in its new configuration and reports all of the nodes and edges listed in workspace cells with non-empty intersection with the obstacle's bounding box. The validity of these nodes and edges is then checked using the collision detection function.



This method offers a trade-off between pre-processing time and update time. The pre-processing phase can be somewhat expensive if the chosen cell size is small due to the well known exponential increase in complexity incurred by grid data structures. On the other hand if the cells are large that will result in many unnecessary and expensive collision detection calls that will be performed as obstacles intersect the grid cell more often during updates, and each such grid cell is associated with a prohibitively long list of affected nodes and edges.

Because of this we report at least two instances of the grid method for every experiment in order to demonstrate that the discrepancy between SPITE and the grid method is not an artifact of the parameters used. 


\subsubsection{\LazyPRM Implementation}
The first single query method we compare against is \LazyPRM\cite{bk-ppulp-00}. We test \LazyPRM by giving it the same roadmap used by our
data structure as an argument. \LazyPRM connects the start and goal
configurations to the roadmap graph, finds a path in the graph,
validates its vertices, and only validates its edges if all previous
steps were successful. The path validation process uses hierarchical
validation, meaning, first attempts to validate the path at a coarse
resolution (one in every 27 intermediates of an edge), and only if the
entire path is valid at that resolution moves forward to a finer
resolution (one in 16) and finally completely validates the path.

Since this implementation of \LazyPRM
first validates the vertices of a path before launching an expensive edge
validation, we ensure that at least in the mobile robot experiment the
validity of the nodes along a path are a perfect indication of the validity
of its edges, thus maximizing the efficiency of \LazyPRM as much as possible.
Since the roadmaps used for the manipulators were randomly created we did
not guarantee this property in those experiments.

\subsubsection{\RRT Implementation}
The second method we compare against is \RRT\cite{l-rrtntpp-1998}. Our implementation is
standard. We use minimum and maximum extension distances to be
$0.1$ and $4.0$ respectively. Out of all the methods we use in this
experiment \RRT is the only one which does not use an initial roadmap
since it performs pathfinding directly in $C$-space. All other methods use
a roadmap as an argument and perform path finding only on that roadmap.

\paragraph{General implementation details} All experiments were run on a desktop computer with an Intel Core i7-3770 at 3.40GHz, 16 GB of RAM. The Parasol Planning Library (PPL) implementations were used for all \SBMP functions and algorithms. 

\subsection{Dynamic Roadmap Updates}
\label{sec:update:experiments}

In the first experiment, we test our graph update method against the grid method
and a brute-force method. Our method and the grid method compute a set of nodes and
edges and then validate only that set using fine-grained collision detection. The brute-force method re-validates every
node and edge in the graph with no regard to the workspace changes that took place.
Note that all methods use the same collision detection algorithm which only performs
collision detection against the subset of obstacles whose position has changed.

\begin{figure}[t]
\centering
    \includegraphics[width=0.55\linewidth]{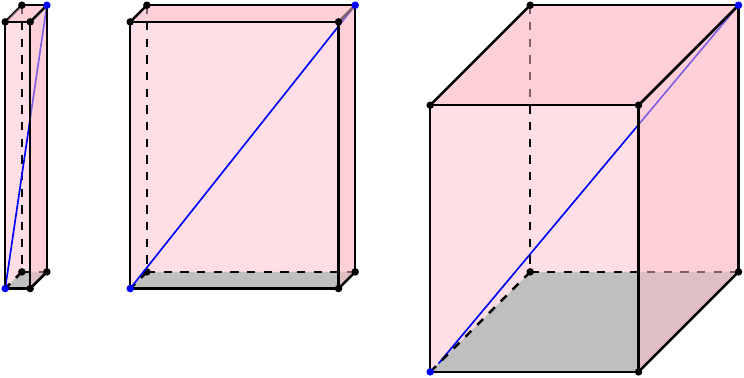}
    \caption{Different orientations of similar line segment obstacles (in blue) and
    their corresponding axis-aligned bounding boxes. }
    \label{fig:update:environment:bounding:boxes}
\end{figure}

The experiment environment is a cube $[-16,16]^3$ with a single axis-aligned
rectangular prism obstacle. The robot is a simple 6\DOF rectangular prism.
Experiments are run with nine different shapes and sizes of the rectangular prism
obstacle, which results in different queries to the grid data structure and
the cigar tree. We use rectangular obstacles of varying sizes and shapes because
it mimics the bounding boxes of differently shaped obstacles. Both data structures,
i.e. the cigar-tree and the grid, use the axis-aligned bounding box of the obstacle
when computing the intersected regions (tree leaves and grid cells respectively).
For example, the update operation required for an arbitrarily oriented long and thin
obstacle $o$ will result in a data structure query which could be a long and thin
axis-aligned box if $o$'s orientation is close to parallel to one of the
axes and perpendicular to all other axes, but could also be a ``slice'' of
the environment or even a large cube if the orientation is diagonal in
some or all of the dimensions. See Figure~\ref{fig:update:environment:bounding:boxes}
for a visualization of this.

Three sets of experiments are performed with cube-shaped obstacles of side lengths
$2, 5,$ and $20$, and six sets experiments are performed with prisms of dimensions
$[20\times20\times1], [20\times1\times1], [20\times20\times2], [20\times2\times2],
[20\times5\times5],$ and $ [20\times20\times5]$.

We created a roadmap with 1000 nodes and $\sim$5000 edges using a uniformly
sampling \PRM algorithm where each node is connected to its 6 nearest
neighbors. Then we constructed our data structures with the roadmap as input. Both
our data structure and the grid method have a pre-processing phase where the graph
nodes and edges are mapped to workspace objects (cigars and grid cells
respectively) and checked for validity. Notice the time required for this
pre-processing step is not included in the average comparison time, but is included
as part of the reported results. We perform 100 random location changes for each
obstacle size, each followed by a roadmap update operation with the change to the
obstacle location as an input. We measure and compare the time required to perform
an update.

The results are shown in Table~\ref{table:Update:times}.

\newcolumntype{s}{>{\centering\arraybackslash}r}
\newcolumntype{C}{>{\centering\arraybackslash}c}
\begin{table}[t!]
\centering
    \begin{tabular}{|m{1.7cm}|s|s|s|s|s|s|}
    \hline
    \textbf{obstacle} &
      \multicolumn{1}{C|}{\textbf{SPITE(ms)}} &
      \multicolumn{1}{C|}{\textbf{Grid\_1(ms)}} &
      \multicolumn{1}{C|}{\textbf{Grid\_2(ms)}} &
      \multicolumn{1}{C|}{\textbf{Grid\_4(ms)}} &
      \multicolumn{1}{C|}{\textbf{Grid\_8(ms)}} &
      \multicolumn{1}{C|}{\textbf{BF}} \\ \hline
    $1 \times 1 \times 1$   & \cellcolor{green!50} 5.3   & 8.9   & 15.8  & 36.0  & 116.3 & \multirow{8}{*}{$\sim$ \text{4 sec}} \\ \cline{1-6}
    $1 \times 1 \times 1$   & \cellcolor{green!50} 50.3  & 67.0  & 93.9  & 149.8 & 296.9 & \\ \cline{1-6}
    $20 \times 2 \times 2$  & \cellcolor{green!50} 39.7  & 57.2  & 85.2  & 152.6 & 343.1 & \\ \cline{1-6}
    $5 \times 5 \times 5$   & \cellcolor{green!50} 58.0  & 70.7  & 95.7  & 153.4 & 301.4 & \\ \cline{1-6}
    $20 \times 20 \times 1$ & \cellcolor{green!50} 212.2 & 246.9 & 314.3 & 453.0 & 718.8 & \\ \cline{1-6}
    $20 \times 5 \times 5$  & \cellcolor{green!50} 136.7 & 166.1 & 212.7 & 300.9 & 478.3 & \\ \cline{1-6}
    $20 \times 20 \times 2$ & \cellcolor{green!50} 292.1 & 337.9 & 417.5 & 577.8 & 872.9 & \\ \cline{1-6}
    $20 \times 20 \times 5$ & \cellcolor{green!50} 422.4 & 465.6 & 538.7 & 708.6 & 974.9 & \\ \cline{1-6}
    $20 \times 20 \times 20$& \cellcolor{green!50} 610.7 & 778.9 & 773.8 & 924.4 & 1126.8& \\ \hline
    Pre-processing time     & \cellcolor{green!25} 2 min & 29 hrs& 4 hrs & 34 min& 4 min & \cellcolor{green!50} -\\ \hline
    \end{tabular}
    \caption{Comparison of dynamic roadmap updates using SPITE and grid with
    different cell sizes. BF stands for brute force, and Grid\_i refers to the grid
    method with cells of side length $i$. All results for which a time unit is not
    disclosed are in milliseconds. Best result in every row is highlighted green, the last row has lighter green for best pre-processing time out of methods with such a runtime component.}
    \label{table:Update:times}
\end{table}



\subsection{Motion Planning Queries}
\label{sec:motion:planning:experiments}

A natural way of using a data structure such as ours is using it to perform
obstacle update operations on a roadmap when obstacles change
position, and use that roadmap to answer motion planning queries
as they are given. Another way to handle environment changes between motion
planning queries is to not store any $C$-space information and solve motion planning queries using a single query algorithm. 
In this set of experiments, we evaluate this latter option against our data structure and the grid method  by comparing the run-time of single query algorithms against the combined runtime of a data structure update method and a motion planning query using the roadmap.

\begin{figure}[t]
    \centering
    \hfill
    \begin{subfigure}{0.49\linewidth}
        \includegraphics[width=\linewidth]%
        {\si{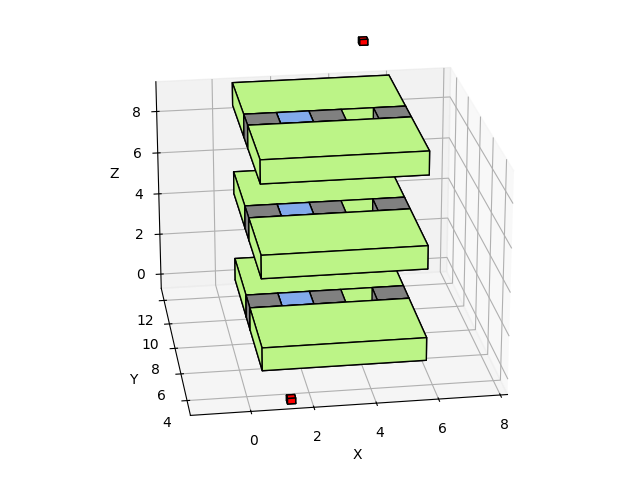}}
        \caption{The starting position of every obstacle in the
           environment\vspace{0.35cm}}
        \figlab{motion:planning:env:1}
    \end{subfigure}
    \hfill
    \begin{subfigure}{0.49\linewidth}
        \includegraphics[width=\linewidth]%
        {\si{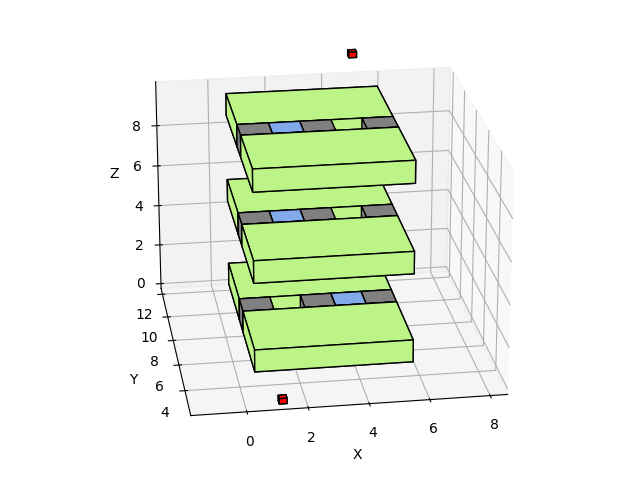}}
        \caption{A random movement of the blue obstacles. Only the lowest
           obstacle switched to other position.}
        \figlab{motion:planning:env:2}
    \end{subfigure}

    \caption{An illustration of the environment used for the experiments described in Section \ref{sec:motion:planning:experiments}. Two configurations of the translational cube robot are seen in red. In every Iteration of the experiment each of the blue obstacles changed its position with probability $1/2$. Note that the bounding box is not illustrated for the sake of clarity. }
       
    
    \figlab{motion:planning:env}
\end{figure}


\subsubsection{Mobile Robot Experiment}
\label{sec:mobile:robot:mp:experiment}
We run the experiment on the environment shown in
Figure \ref{fig:motion:planning:env}, which contains three walls, each with two
possible locations of a narrow passage. The experiment consists of 1000 iterations, in each of which we change the location of the passage in each wall with
probability $1/2$. Our robot is a small translational cube of
side-length $1/4\th$ that of the passage. Note that using a 6\DOF
robot would increase the search space of \RRT while not
changing anything about the runtime of the other methods which use a fixed size roadmap. 
For each iteration, we generate random start and goal queries in 
opposite sides of the environment, requiring each query to find a 
path through all three narrow passages. All methods except \RRT are provided
the same $C$-space roadmap. This roadmap is an induced subgraph of the 3D
graph with resolution (i.e. edge length $1/4$) on $\sim 3400$ nodes, and is guaranteed to contain a valid path through all three walls
regardless of obstacle locations and, therefore, from any sampled start
to any sampled goal configurations.

\begin{table}[t!]
    \centering
    \begin{tabular}{|l|r|r|r|r|} \hline
      Query Method&Update(s)&Query (s)&Total&Pre-processing (min)\\ \hline
      \PRM + SPITE & 0.13 \cellcolor{green!50} & 0.12 \cellcolor{green!50} & 0.25 \cellcolor{green!50} & 0.95 \cellcolor{green!25}\\ \hline
      \PRM + Grid\_1    & 0.14 & 0.15 & 0.28 & 40.3 \\ \hline
      \PRM + Grid\_2    & 0.39 & 0.14 & 0.52 & 7.6\\ \hline
      \LazyPRM     & - & 1.84 & 1.84 & - \cellcolor{green!50}\\ \hline
      \RRT         & - & 10.68 & 10.68 & - \cellcolor{green!50}\\ \hline
    \end{tabular}
    \caption{Comparison of graph-based motion planning algorithms augmented for modified environments and single-query motion planning algorithms in a mobile robot scenario. Grid\_i indicates the grid method with cells of side length $i$, and the best results are shown in green. Note that the best pre-processing time out of the methods with a pre-processing phase is highlighted in lighter green.}
    \label{table:motion:planning}
\end{table}

The results of the experiment are presented in Table~\ref{table:motion:planning}.


\subsubsection{Manipulator Experiment}
\label{sec:manipulator:mp:experiment}

For this experiment we use a simulation of a 5\DOF UR5e in a shelf environment shown in Figure~\ref{fig:ur5}. The two-story shelf contains two obstacles, a spray can and a box which are located at the center of either the top or bottom shelf with probability $1/2$ each. The simulated environment size is $6\times 6\times 4$ meters, though out of these the volume reachable by the robot is much smaller, and the physical setup is contained in a $2 \times\ 2 \times 1$ meter region. All algorithms except for \RRT were provided with a roadmap containing 1000 nodes that was guaranteed to contain a path regardless of the obstacles' positions. We performed 100 iterations using a single start $s$ and goal $t$ pair, where the robot's endeffector is located at the back right corner of the upper shelf in configuration $s$, and in the back left corner in configuration $t$. 
In this experiment we have capped the runtime of \RRT at 60 seconds.

The paths produced in this experiment specifically for the instances where both objects were placed in the same shelf were validated using the physical setup shown in Figure~\ref{fig:ur5} and in the videos provided as supplemental material.


\begin{table}[t!]
    \centering
    \begin{tabular}{|l|r|r|r|r|} \hline
      Query Method&Update(s)&Query (s)&Total&Pre-processing (min)\\ \hline
      \PRM + SPITE & 0.199 \cellcolor{green!50}& 0.028 \cellcolor{green!50}& 0.227 \cellcolor{green!50}& 27.6\\ \hline
      \PRM + Grid\_0.5    & 0.773 & 0.029 \cellcolor{green!50}& 0.802 & 89.9 \\ \hline
      \PRM + Grid\_1    & 9.310 & 0.165 & 9.475 & 12.3 \cellcolor{green!25}\\ \hline
      \LazyPRM     & - & 0.349 & 0.349 & - \cellcolor{green!50}\\ \hline
      \RRT         & - & *0.016 \cellcolor{red!25}& *0.016 \cellcolor{red!25}& - \cellcolor{green!50}\\ \hline
    \end{tabular}
    \caption{Comparison of graph-based motion planning algorithms augmented for modified environments and single-query motion planning algorithms for a UR5e in a shelf environment. Grid\_i indicates the grid method with cells of side length $i$ meters, and the best results are shown in green. Note that the best pre-processing time out of the methods with a pre-processing phase is highlighted in lighter green, and runtimes hiding a success rate of only 53 percent are highlighted red.}
    
    \label{table:ur5}
\end{table}
\footnotetext[1]{53\% success rate}
The results of the experiment are presented in Table~\ref{table:ur5}. Notice that \RRT had a success rate of 53 percent.

\subsection{Discussion}

The first experiment, described in Section \ref{sec:update:experiments}, demonstrates that for many ``reasonably'' shaped obstacles SPITE outperforms the grid method in update times while requiring a fraction of the preprocessing time. As the size of the obstacle increases to a diameter more than half that of the environment we see in
Table~\ref{table:Update:times} a clear increase in update times due to the sheer number of nodes and edges which are invalidated and re-validated at every iteration.


The most glaring detail about this experiment is the exponential increase in pre-processing time that plagues the grid method, with a factor of close to the expected $2^d$ factor in the runtime with every refinement of the grid that is translated to diminishing returns in update performance, resulting in a 29 hours pre-processing time required to construct a data structure with $32^3$ cells in order to compete with our method. This is by far the largest environment of all of the experiments and thus the one where this effect can be most easily seen.

The table also shows us that while there is no clear trend in the relative difference between update times of SPITE and the grid method, there is an increase in absolute difference as obstacles increase in size.
Note that this result carries over to the case of multiple updated obstacles as all methods will sequentially perform the updates for each moved obstacle.

The second and third experiments, described in Section \ref{sec:motion:planning:experiments}, showcase the contribution of these faster updates when performing motion planning queries.

In the mobile robot experiment, described in the first part of Section \ref{sec:mobile:robot:mp:experiment} and summarized in Table~\ref{table:motion:planning}, we see that unlike the dynamic roadmaps, \RRT and \LazyPRM are both affected by the distance between the start and goal configurations, as it directly increases the number of collision checks they need to perform, and all of the methods that utilize a roadmap benefit greatly from circumventing the need to perform pathfinding in $C$-space. 

The small modifications which are due to the small obstacles and a small and fat robot result in smaller scale updates and thus a lead to mild absolute update time differences. 


In the last experiment, described in the second part of Section \ref{sec:manipulator:mp:experiment} and summarized in in Table~\ref{table:ur5}, we see that as the robot becomes more complicated, e.g. composed of more bodies, some of which are not fat, and as the obstacles occupy more of the reachable volume required to perform the motion planning task, the runtime difference is strongly pronounced with our method requiring $\sim 60$ percent less runtime than the grid method while requiring significantly less pre-processing time. Notice that in this more complicated scenario even a very coarse grid requires a substantial preprocessing time. We believe that the extremely sharp decrease in query time is a result of large fractions  of the edges in the roadmap intersecting a few grid cells with side length 1.

\LazyPRM does indeed perform much better due to the relatively short distances, a fact which is not enough to compensate for the difficulty of the pathfinding task when the top shelf contains one of the objects thus causing \RRT to not be able to complete the task within the allotted 60 seconds. 


\section{Conclusion}

In this paper, we introduced the SPITE supplementary method for dynamic roadmaps and an approach to 3D volume approximation with a matching hierarchical collision checking method for nodes and edges in $C$-space graphs. We found our method outperforms the only previously known method in both update and pre-processing times, and, when used in the context of motion planning queries can lead to faster motion planning queries for mobile robots and manipulators in modified environments than single-query algorithms.

For future work, we plan to focus on the 3D swept volume approximation, adding lower-bound approximations and using oriented bounding boxes instead of cigars, and add lazy planning heuristics to our method.



\bibliographystyle{plain}
\bibliography{main.bib}

\end{document}